\documentclass{article}

\usepackage[final]{nips_2017}

\usepackage[utf8]{inputenc} 
\usepackage[T1]{fontenc}    
\usepackage[hidelinks]{hyperref}       
\usepackage{url}            
\usepackage{booktabs}       
\usepackage{amsfonts}       
\usepackage{nicefrac}       
\usepackage{microtype}      
\usepackage{graphicx}
\usepackage{verbatim}

\usepackage{multicol} 
\usepackage{caption}
\usepackage{listings}

\usepackage{EvolvingAIcommenting}
\newif\ifcomments

\commentstrue

\ifcomments
\newcommand{\comments}[1]{#1}
\else
\newcommand{\comments}[1]{}
\fi

\title{AI-GAs: AI-generating algorithms, an alternate paradigm for producing general artificial intelligence}

\author{
Jeff Clune  \\
  Uber AI Labs, University of Wyoming\\
}

\begin{document}

\maketitle

\begin{abstract}
Perhaps the most ambitious scientific quest in human history is the creation of general artificial intelligence, which roughly means AI that is as smart or smarter than humans. The dominant approach in the machine learning community is to attempt to discover each of the pieces that might be required for intelligence, with the implicit assumption that at some point in the future some group will complete the Herculean task of 
figuring out how to combine all of those pieces into an extremely complex machine. I call this the ``manual AI approach.'' 
This paper describes another exciting path that ultimately may be more successful at producing general AI. It is based on the clear trend from the history of machine learning that hand-designed solutions eventually are replaced by more effective, learned solutions. The idea is to create an AI-generating algorithm (AI-GA), which itself automatically learns how to produce general AI. Three Pillars are essential for the approach: (1) meta-learning architectures, (2) meta-learning the learning algorithms themselves, and (3) generating effective learning environments. While work has begun on the first two pillars, little has been done on the third. Here I argue that either the manual or AI-GA approach could be the first to lead to general AI, and that both are worthwhile scientific endeavors irrespective of which is the fastest path. Because both approaches are roughly equally promising, and because the machine learning community is mostly committed to the engineered AI approach currently, I argue that our community should shift a substantial amount of its research investment to the AI-GA approach. To encourage such research, I describe promising work in each of the Three Pillars. I also discuss the safety and ethical considerations unique to the AI-GA approach. Because it may be the fastest path to general AI and because it is inherently scientifically interesting to understand the conditions in which a simple algorithm can produce general intelligence (as happened on Earth where Darwinian evolution produced human intelligence), I argue that the pursuit of AI-GAs should be considered a new grand challenge of computer science research. 

\end{abstract}

\section{Two approaches to producing general AI: the manual approach vs.\ AI-generating algorithms}

Arguably the most ambitious scientific quest in human history is the creation of general artificial intelligence, which roughly means AI as smart or smarter than humans\footnote{This paper will not attempt to define general intelligence aside from saying it roughly means AI as smart or smarter than humans. Nor does this paper engage in the debate about to what extent such a thing exists. Such terrain is well-trodden without resolution, and is not the focus of this paper.}. The creation of general AI would transform every aspect of society and economic sector. It would also catalyze scientific discovery, leading to unpredictable advances, including further advances in AI itself (the potential ethical consequences of which I discuss in Section \ref{ethics}). 

\subsection{The manual AI approach}
\label{manualApproach}
One approach to producing general AI is via a two-phase process. In Phase One, we discover each of the pieces that might be required for intelligence. In Phase Two, we then figure out how to put all of those pieces together into an extremely complex machine. This path has been followed since the earliest days of AI, such as initial research into systems intended to capture logical thought or process language \cite{russell1995artificial}. 
This is also the path implicitly being taken by the vast majority of the machine learning research community. Most papers introduce or refine a single building block of intelligence, such as convolution \cite{lecun1998gradient}, recurrent gated cells \cite{hochreiter:nc97,cho2014properties, chung2014empirical}, skip connections \cite{he2015deep,srivastava2015highway}, 
attention-mechanisms \cite{xu2015show,devlin2018bert},
activation functions \cite{Goodfellow-et-al-2016,hahnloser2000digital,nair2010rectified,maas2013rectifier,clevert2015fast},
external memory \cite{graves2016hybrid,weston2014memory},
good initializations \cite{glorot2010understanding,Goodfellow-et-al-2016},
normalization schemes \cite{ioffe2015batch,lei2016layer},
hierarchical methods \cite{dayan1993feudal,barto2003recent,hausman2018learning,vezhnevets2017feudal},
capsules \cite{sabour2017dynamic,hinton2011transforming},
unsupervised learning techniques \cite{Goodfellow-et-al-2016,radford2015unsupervised,le2011building,goodfellow2014generative},
value or advantage functions \cite{sutton1998reinforcement}, 
intrinsic motivation \cite{stanton2018deepCS,pathak2017curiosity,oudeyer2009intrinsic,burda2018large}, 
trust regions \cite{schulman2015trust},
auxiliary tasks \cite{jaderberg2016reinforcement}, 
solutions to catastrophic forgetting \cite{kirkpatrick2017overcoming, ellefsen2015neural, zenke2017continual,velez2017diffusion}, and many, many more. Table \ref{tableOfBBs} lists many example building blocks, and this list is far from exhaustive. Seeing the length of this incomplete list raises the following questions: How long will it take to identify the right variant of each building block in this list? How many more essential building blocks exist that we have yet to invent? How long will it take us to discover them all? 

Even if we were able to create effective versions of each of the key building blocks to general AI in a reasonable amount of time, our work would remain unfinished. What is rarely explicitly mentioned is that collecting key building blocks of intelligence is only useful in terms of building general AI if at some point some group takes on the Herculean scientific challenge of figuring out how to combine of all of those pieces into an intelligent machine. Combining multiple, different building blocks is extraordinarily difficult due to (1) the many different ways they can be combined, (2) non-linear interaction effects between modules (e.g.\ a version of a module can work well when combined with some building blocks, but not others), and (3) the scientific and engineering challenges inherent in working with complex systems of dozens or hundreds of interacting parts, each of which is a complex, likely black-box, machine learning module (e.g.\ imagine trying to debug such a machine or identifying which piece needs to be tweaked if things are not working). There have been notable, valiant, successful efforts to combine multiple (e.g.\ around six to thirteen) AI building blocks \cite{jaderberg2018human,hessel2017rainbow}, but such attempts still feature a low number of building blocks vs.\ the dozens or hundreds that may be required. Such efforts are also rare and understandably are unable to assess the contribution of each separate building block, let alone different combinations and/or configurations of each building block. Such knowledge may be necessary, however, for us to engineer extreme levels of intelligence into machines. 

A final challenge with combining dozens or hundreds of separate building blocks is that doing so does not work well within our scientific culture. Current scientific incentives, organizations, and traditions usually reward small teams of scientists to produce a few papers per year. Combining all of the required AI building blocks together instead would likely require an extremely large, dedicated team to work for years or decades in something akin to an Apollo program. While some excellent institutions like DeepMind and OpenAI are large, well-funded, and focused on creating general AI, they still tend to feature smaller teams focused on separate projects and publishing regularly, instead of an organization where all hands on deck are committed to one project, which may be required to engineer AI. Of course, their structures could evolve with time once the building blocks are thought to have been identified. 

To be clear, I believe the manual AI path (1) has a chance of being the fastest path to AI (Section ~\ref{whichIsFastest}) and (2) is worthwhile to pursue even if it is not (Section \ref{bothPathsWorthwhile}). That said, it is important to make its assumptions explicit---that it is a manual engineering approach involving the two phases described above---and to acknowledge how daunting of a challenge it is, for scientific, engineering, and sociological reasons. Making those points is one of the objectives of this manuscript. 
\begin{table}
\caption{A very partial list of the potential building blocks for AI we might need for the manual path to creating AI. When will we discover the right variant of each of these building blocks? How many more are essential, but are yet to be discovered? How will we figure out how to combine them into a complex thinking machine? Seeing even this partial list underscores how difficult it will be to succeed in creating general AI via the manual path. The idea of AI-GAs is to learn each of these building blocks automatically.  The First Pillar, meta-learning architectures, could potentially discover the building blocks (or better alternatives) colored black (solid circles). The Second Pillar, meta-learning learning algorithms, could potentially learn the red building blocks (diamonds). The Third Pillar, generating effective learning environments, could learn things like the blue building blocks (daggers). \new{Note that placing elements into only one of the pillars is challenging. For example, some solutions (e.g. spatial transformers \cite{jaderberg2015spatial}) could be implemented via an architectural solution (Pillar 1) or be learned to be performed within a recurrent neural network without any architectural changes (Pillar 2). Additionally, if certain learning environments encourage the learning of performing a function or skill (e.g. transforming inputs spatially), then it could also be listed in Pillar 3. Thus, the categorization is to give the rough impression that each of the building blocks in this table \emph{could be learned} in the AI-GA framework by \emph{at least one} of the pillars, rather than to commit to in which pillar it is best learned. Additionally, to achieve both breadth and depth, in some cases I include a broad class (e.g. unsupervised learning) and instances of techniques within that class (e.g. generative adversarial networks \cite{goodfellow2014generative}). An AI-GA researcher might need (or choose) to provide some of these elements, especially the "materials" neural networks are made of (e.g. neuromodulatory neurons and regular neurons) and other AI-GA hyperparameters (Sections \ref{Pillar2-learningAlgos} and \ref{discussion}). }}
\label{tableOfBBs}
   \begin{multicols}{3}
    \begin{itemize}
    \item convolution
    \item pooling
\item activation functions
\item gradient-friendly architectures (e.g.\ skip connections)
\item gated recurrent units (e.g.\ LSTMs \& GRUs)
\item structural organization (regularity, modularity, hierarchy) 
\item spatial tranformers
\item capsules
\item batch/layer norm
\item experience replay buffers
\item writeable memory (e.g. memory networks, neural Turing machines)
\item recurrent layers
\item multi-modal fusion
\textcolor{maroon}
{
\item [$\diamond$] learned losses (e.g.\ evolved policy gradients)
\item [$\diamond$] hierarchical RL, options
\item [$\diamond$] intelligent exploration (e.g.\ via intrinsic motivation)
\item [$\diamond$] good initializations (Xavier, MAML, etc.)
\item [$\diamond$] universal value function approximators
\item [$\diamond$] auxiliary tasks (predicting states, autoencoding, predicting rewards, etc.) 
\item [$\diamond$] catastrophic forgetting solutions 
\item [$\diamond$] efficient continual learning 
\item [$\diamond$] Dyna
\item [$\diamond$] variance reduction techniques 
\item [$\diamond$] good hyperparameters 
\item [$\diamond$] value functions, action-value functions, advantage functions
\item [$\diamond$] models of the world
\item [$\diamond$] planning algorithms
\item [$\diamond$] backprop through time
\item [$\diamond$] causal reasoning
\item [$\diamond$] trust regions
\item [$\diamond$] Bayesian learning
\item [$\diamond$] neuromodulation
\item [$\diamond$] Hebbian learning
\item [$\diamond$] active learning
\item [$\diamond$] probabilistic models
\item [$\diamond$] generative adversarial networks
\item [$\diamond$] time-contrastive, object-contrastive losses
\item [$\diamond$] audio-visual correspondence
\item [$\diamond$] domain confusion
\item [$\diamond$] entropy bonuses 
\item [$\diamond$] regularization 
\item [$\diamond$] distance metrics
\item [$\diamond$] hindsight experience replay
\item [$\diamond$] attention mechanisms
\item [$\diamond$] complex optimizers (Adam, RMSprop, etc.)
\item [$\diamond$] learning rates \& schedules
\item [$\diamond$] discount factors
\item [$\diamond$] eligibility traces
\item [$\diamond$] theory of mind
\item [$\diamond$] self-modeling
\item [$\diamond$] unsupervised learning
}
\textcolor{blue}
{
\item [$\dagger$] co-evolution / self-play
\item [$\dagger$] staging / shaping 
\item [$\dagger$] curriculum learning
\item [$\dagger$] communication / language
\item [$\dagger$] multi-agent interaction
\item [$\dagger$] diverse environmental challenges
\item [$\dagger$] environmental hyperparameters (e.g.\ episode length)
}
    \end{itemize}
    \end{multicols}
\end{table}

\subsection{A different approach: AI-generating algorithms (AI-GAs)}
\label{AIGAs-aDifferentApproach}

There is another exciting path that ultimately may be more successful at producing general AI: the idea is to learn as much as possible in an automated fashion, involving an \emph{AI-generating algorithm} (AI-GA) that bootstraps itself up from scratch to produce general AI. As argued below, this approach also could prove to be the fastest path to AI, and is interesting to pursue even if it is not. 

One motivation for a learn-as-much-as-possible approach is the history of machine learning. There is a repeating theme in machine learning (ML) and AI research. When we want to create an intelligent system, we as a community often first try to hand-design it, meaning we attempt to program it ourselves. Once we realize that is too hard, we then try to hand-code some components of the system and have machine learning figure out how to best use those hand-designed components to solve a task. Ultimately, we realize that with sufficient data and computation, we can learn the entire system. The fully learned system often performs better. It also is more easily applied to new challenges (i.e.\ it is a more general solution). 

The clearest example of this trend is in computer vision. First we tried to hand-code computational vision systems in the early decades of AI research and found that they did not work well \cite{Szeliski2010ComputerVisionTextbook}. We then switched to a hybrid approach wherein we manually designed features (e.g.\ edge-, texture-, object-, and other feature-detectors, including the famous HOG \cite{dalal2005histograms} and SIFT \cite{lowe1999object} feature sets) and then had machine learning algorithms automatically learn the best way to use those hand-coded features to solve a particular problem \cite{Szeliski2010ComputerVisionTextbook}. Starting around 2012 \cite{krizhevsky2012imagenet}, we realized that we had sufficient computation and data to learn entire vision systems without injecting hand-designed features. Doing so has improved performance and such systems are very general: they can be applied to a wide variety of vision tasks and learn appropriate features for each. This same phenomenon has occurred in other modes of machine learning, such as sound (e.g.\ speech-to-text systems) \cite{hinton2012deep}, touch \cite{gao2016deep}, natural language understanding and translation \cite{devlin2018bert,bahdanau2014neural,sutskever2014sequence}, and many more \cite{Goodfellow-et-al-2016}. 

This trend also has occurred in other areas of machine learning. For decades, but with especially intense effort since 2012, we as a community have been hand-designing increasingly powerful neural network architectures \cite{he2015deep,huang:arxiv16,szegedy:arxiv16,Goodfellow-et-al-2016}. Increasingly, however, researchers are searching for deep neural network architectures automatically with machine learning algorithms \cite{real:arxiv18,rawal:arxiv18,miikkulainen:arxiv17,zoph2016neural}. Despite all of the considerable effort that has gone into designing architectures for image classification manually, an architecture search algorithm produced the state of the art on the CIFAR benchmark at the time of its publication \cite{real2018regularized}. The state of the art on the extremely popular and over-optimized ImageNET benchmark also belongs not to a human-designed architecture, but to one produced by an architecture search algorithm \cite{real2018regularized}. Additionally, while most hyperparameters have been found by hand throughout the history of machine learning, they are increasingly learned \cite{snoek2012practical,jaderberg2017population}. Additionally, there is a recent increase in meta-learning the learning algorithms themselves \cite{duan2016rl,wang2016learning,ellefsen2015neural, velez2017diffusion,soltoggio2018born} (discussed more in Section \ref{Pillar2-learningAlgos}). 

I am amongst those that believe that this trend of learned pipelines replacing hand-designed ones will continue, but increasingly it will apply to the machinery of machine learning itself. That opens the tantalizing prospect that we might be able to create an algorithm that learns everything required to produce general AI, and thus produce an AI-GA. That would obviate the need to discover, refine, and combine the individual building blocks of an intelligent machine (e.g.\ those in Table \ref{tableOfBBs} and the many yet to be discovered). 

A second motivation for a learn-as-much-as-possible approach is that it is the only approach that we are certain can work. That is because it already has. On Earth, a remarkably unintelligent algorithm in Darwinian evolution was paired with the correct conditions and bootstrapped itself up from the simplest replicators to ultimately producing the human mind. Darwinian evolution thus is the first general-intelligence-generating algorithm, providing an existence proof that the concept of general-intelligence-generating algorithms can work. Natural evolution required unfathomable amounts of computation, however. An open scientific question is how we can create abstractions of what drove the success of natural evolution in order to produce AI-GAs that can work within the computation that we will have in the coming years. 

\subsection{The Three Pillars required to produce an AI-GA}

There are Three Pillars that, if we could get them to work well, should produce general AI. Thus, research into AI-GAs should focus on improving and combining them. The Three Pillars are (1) meta-learning architectures, (2) meta-learning the learning algorithms themselves (often just called meta-learning), and (3) automatically generating effective learning environments. Section ~\ref{researchIntoTheThreePillars} describes each pillar and what research into it might look like. Briefly, research into the First and Second Pillars is decades old, but both have seen a surge of recent interest and activity. There has been very little research into the Third Pillar, although we recently published our first step in this direction \cite{wang2019paired}. Thus the Third Pillar is the least-studied and least-understood. I consider it the hardest. I predict that more history-making discoveries await in this pillar than the other two, making it an exciting, fruitful area of research. 

\subsection{AI-GAs: A new grand challenge for AI research}
\label{AI-GAIsAGrandChallenge}

AI-GAs may prove to be the fastest path to general AI. However, even if they are not, they are worth pursuing anyway. It is intrinsically scientifically worthwhile to attempt to answer the question of how to create a set of simple conditions and a simple algorithm that can bootstrap itself from simplicity to produce general intelligence. Just such an event happened on Earth, where the extremely simple algorithm of Darwinian evolution ultimately produced human intelligence. Thus, one reason that creating AI-GAs is beneficial is that doing so would shed light on the origins of our own intelligence.

AI-GAs would also teach us about the origins of intelligence generally, including elsewhere in the universe. They could, for example, teach us which conditions are necessary, sufficient, and catalyzing for intelligence to arise. They could inform us as to how likely intelligence is to emerge when the sufficient conditions are present, and how often general intelligence emerges after narrow intelligence does. 

Presumably different instantiations of AI-GAs (either different runs of the same AI-GA or different types of AI-GAs) would lead to different kinds of intelligence, including different, alien cultures. AI-GAs would likely produce a much wider diversity of intelligent beings than the manual path to creating AI, because the manual path would be limited by our imagination, scientific understanding, and creativity. We could even create AI-GAs that specifically attempt to create different types of general AI than have already been produced. AI-GAs would thus better allow us to study and understand the space of possible intelligences, shedding light on the different ways intelligent life might think about the world.

The creation of AI-GAs would enable us to perform the ultimate form of cultural travel, allowing us to interact with, and learn from, wildly different intelligent civilizations. Having a diversity of minds could also catalyze even more scientific discovery than the production of a single general AI would. We would also want to reverse engineer each of these minds for the same reasons we study neuroscience in animals, including humans: because we are curious and want to understand intelligence. We and others have been conducting such `AI Neuroscience' for years now \cite{nguyen2015deep,nguyen2016multifaceted,nguyen2017plug,nguyen2016synthesizing,yosinski2015understanding,li2016convergent, yosinski2014transferable,szegedy2013intriguing,mahendran2014understanding,zeiler2014visualizing,li2015visualizing,greydanus2017visualizing}, but the work is just beginning. 
John Maynard Smith wrote the following about evolutionary systems: ``So far, we have been able to study only one evolving system, and we cannot wait for interstellar flight to provide us with a second. If we want to discover generalizations about evolving systems, we will have to look at artificial ones.'' \cite{Maynard-Smith:1992aa} The same is true regarding intelligence. 

Additionally, explained in Section \ref{Pillar3learningEnvs}, the manner in which AI-GA is likely to produce intelligence will involve the creation of a wide diversity of learning environments, which could include novel virtual worlds, artifacts, riddles, puzzles, and other challenges that themselves will likely be intrinsically interesting and valuable. Many of those benefits will begin to accrue in the short-term with AI-GA research, so AI-GAs will provide short-term value even if the long-term goals remain elusive. 
AI-GAs are also worthwhile to pursue because they inform our attempts to create open-ended algorithms (those that endlessly innovate) \cite{stanley:open}. As described in Section \ref{AIGAs-aDifferentApproach}, we should also invest in AI-GA research because it might prove to be the fastest way to produce general AI. 

For all these reasons, I argue that the creation of AI-GAs should be considered its own independent scientific grand challenge. 

\subsection{Both the manual and AI-GA paths are worth pursuing, irrespective of which is more likely to be the fastest path to general AI}
\label{bothPathsWorthwhile}

Both the manual and AI-GA paths to creating general AI should be pursued, regardless of which one is considered the most likely to succeed first. The clearest argument for that is that even if one succeeds, we would still want to pursue the other. To see why, consider the possibility of each being the first to produce general AI. 

If the manual path wins, we would still be interested in creating AI-GAs for all the reasons just provided in Section \ref{AI-GAIsAGrandChallenge}. If the AI-GA path wins, we would still be interested in creating general AI via the manual path. That is because it is likely that an AI-GA would produce a complex machine whose inner workings we do not understand. As Richard Feynman said, ``What I cannot create, I do not understand.'' We understand by building, and one great way to understand intelligence is to be able to take it apart, rewire components, and learn how to put it together piece by piece. AI-GAs could actually catalyze the manual path because it would produce a large diversity of general AIs, perhaps making it easy for us to recognize what is necessary, variable, and incidental in the creation of intelligent machines.  Additionally, each path would catalyze the other because the general AI produced by either could be used to accelerate scientific progress in the other path. 

For all of these reasons, I consider the manual and AI-GA paths to creating general AI to be separate scientific grand challenges each worthy of substantial scientific investment. However, because the current machine learning community is mostly committed to the manual path, I advocate that we should shift investment to the AI-GA path to pursue both promising paths to general AI. 

\subsection{Which path is more likely to produce general AI first?}
\label{whichIsFastest}

There are good reasons why either of the two paths might be the fastest way to produce general AI. However, on balance, after weighing all of the arguments presented in this manuscript, I believe it more likely that the AI-GA path will produce general AI first. That said, I have high uncertainty about that predication and think that either could get there first. 

There are a few main reasons why I think the AI-GA path is more likely. Initially, it scales well with the exponential compute we can expect in the years and decades to come. While AI-GAs will require extraordinary amounts of computation by today's standards, they have the nice property that they will be able to consume that compute easily once it is available. As Richard Sutton has recently pointed out, history has shown that the algorithms that tend to win over the long haul are simple ones that can take advantage of massive amounts of computing \cite{sutton2019bitter}. Additionally, as noted above, history teaches us that learned pipelines eventually surpass ones that are entirely or partially hand-created. Further, the AI-GA path enables small teams to begin working on algorithms that might produce general AI right now. In contrast, as noted above, the manual path eventually requires a switch to the Herculean Phase 2 task of putting together all of the building blocks of intelligence. 
If such a combination requires a very large team dedicated for years (something akin to the Apollo program), then major changes in the social organization of scientific effort must take place before work can begin on anything that has a chance at producing an intelligent machine via the manual path. That work must also be undertaken after, or concurrently with, discovering all of the building blocks of intelligence (Phase 1 of the manual path). 
Additional reasons why the AI-GA path might win are given in Sections \ref{manualApproach} and \ref{AIGAs-aDifferentApproach}.
Overall, it is likely that in the near to middle term, the manual path will produce higher-performing AI systems that are of more use to society, just as HOG and SIFT produced better computer vision systems in the pre-2012 era. However, just as with computer vision, it is likely that fully learned systems will eventually surpass and then rapidly far exceed the capabilities produced by the manual path. 

\section{Research into the Three Pillars}
\label{researchIntoTheThreePillars}

Researching how to create an AI-GA is a very different research agenda from the manual path. Instead of attempting to identify and create the individual building blocks of an intelligent machine (e.g.\ better memory modules, hierarchical controllers, optimizers, and transformers), AI-GAs attempt to learn all of these types of building blocks. That does not mean it is not without its own challenging research problems. It is just that the research problems are different and require focusing on different questions. 

This section describes each of AI-GA's Three Pillars, including prior work in them and a sketch of what future research in them might look like. 

\subsection{The First Pillar: Meta-learning Architectures}

The First Pillar is the most studied and well-known, and for that reason I will spend little time surveying it. The architecture of a neural network plays a key role in both the function a trained neural network can perform and how easy it is to train that network \cite{he2015deep,huang:arxiv16,szegedy:arxiv16,Goodfellow-et-al-2016}. As mentioned above, the majority of work into training neural networks with backpropagation has involved humans hand-designing architectures, including innovations such as convolution \cite{lecun1998gradient}, LSTMs or other gated recurrent units \cite{hochreiter:nc97,cho2014properties, chung2014empirical}, highway networks and residual networks \cite{he2015deep,srivastava2015highway}, etc. However, there is also a long history in AI of creating \emph{algorithms} to search for high-performing neural network architectures \cite{lecun1990optimal,stanley2002evolving,gruau1994automatic,yao1999evolving,harp1990designing}. Starting recently, such architecture search algorithms are being tested with large-scale compute for deep neural network architectures, where they are sometimes finding better architectures than those that humans have designed. For example, automatically discovered architectures improved performance even on the classic benchmark of CIFAR and the challenging benchmark of ImageNET \cite{kandasamy2018neural,real:arxiv18,rawal:arxiv18,miikkulainen:arxiv17,zoph2016neural}, both of which had already been highly optimized by armies of human scientists. We can thus reasonably assume the trend will continue and that searching for architectures will replace the hand-designing of architectures in the years and decades to come. Architecture search is expensive, but so was backpropagation in deep neural networks in the year 2010. Eventually the computation will exist to make it routine and likely the best option. Additionally, there are techniques that can speed it up, such as surrogate models \cite{liu2018progressive,klein2016learning,baker2017accelerating,rawal:arxiv18} or Generative Teaching Networks \cite{petroskisuch2019GTN}, and new techniques will be invented to make it faster still. 

It is clear that animal brains, including our own, are not large, homogenous, fully connected neural networks. Instead, their architectures consist of many heterogenous modules wired together in a particular way, with the entire design carefully sculpted by evolution \cite{striedter2005principles}. It is daunting to consider creating such complex architectures by hand. With exponentially more compute, we instead can search for them with an outer loop learning algorithm that produces powerful, sample-efficient architectures primed for learning. Since the outer-loop algorithm produces architectures that are better at learning, people often call architecture search algorithms `meta-learning algorithms', because they learn how to produce better learners. Another notion of meta-learning (learning better learning algorithms) is described in the next section. 

Much research needs to be done to enable an effective architecture search algorithm. That includes research into encodings (aka representations) \cite{stanley2007compositional,stanley2009hypercube,stanley2003taxonomy,zoph2016neural,gruau1994automatic}, search operators, and the production of architectures that are regular \cite{stanley2009hypercube, clune2011performance,huizinga2014evolving}, modular \cite{clune2013originModularity, huizinga2014evolving}, and hierarchical \cite{mengistu2016evolutionary}. There is the chance for such architecture search algorithms to discover architectural innovations like those colored black in Table \ref{tableOfBBs}. Research in this space is taking off at the moment and will be useful whether the search is for a domain-specific neural network solution or for producing an AI-GA. 

\subsection{The Second Pillar: Meta-learning the learning algorithms}
\label{Pillar2-learningAlgos}

Because history teaches us we should learn as much as possible, that lesson should apply to the learning algorithms themselves. Doing so could improve the ceiling for what neural networks are able to learn about each task, their sample efficiency, their ability to generalize, and their ability to continuously learn many tasks. Work in this field is often called simply `meta-learning' and dates back decades \cite{hochreiter2001learning,cotter1990fixed,soltoggio2018born,schmidhuber1987evolutionary,ellefsen2015neural, velez2017diffusion,finn2017model,wang2016learning,duan2016rl,schmidhuber2015deep,andrychowicz2016learning,ravi2016optimization,li2016learning}. The idea is to `learn to learn' such that an agent gets better at learning a new task sampled from a distribution of tasks over the course of training. Instead of learning a specific task, the idea is to get better at learning on any new task by becoming a better learner. There has been a recent surge of interest in this area, with work falling into roughly two families. 

The first family, made famous by the Model-Agnostic Meta Learning (MAML) algorithm \cite{finn2017model} involves pairing a learner (a neural network with a set of weights $\theta$) with a predefined, fixed learning algorithm---such as stochastic gradient descent (SGD)---to solve a task selected from a distribution of tasks as fast as possible. The job of the learning algorithm is to produce an initial set of weights that can rapidly learn any task from the distribution. To do so, MAML differentiates through the learning process to find a set of initial weights that are quickly able to learn a new task when taking gradient steps via SGD in that new task. It has been shown that theoretically this paradigm can encode any learning algorithm \cite{finn2017meta}. 

The second family involves training a recurrent neural network (RNN) in a context that requires learning and relies on the activations with the neural network to implement a learning algorithm. Because an RNN is a Turing-complete universal computer \cite{siegelmann1995computational}, it too can encode any possible learning algorithm. The weights of the RNN are trained via an outer loop meta-learning algorithm. This approach has been well-studied in a supervised learning context \cite{hochreiter2001learning,cotter1990fixed}.
As applied to reinforcement learning, it was concurrently introduced by two papers \cite{wang2016learning,duan2016rl}. It is intuitively appealing because it resembles what happened on Earth, where an outer loop optimization algorithm (evolution) created animal brains, which are effective, sample-efficient learners. Note, however, that evolution is not required: many algorithms can be used for the outer-loop learning algorithm, such as policy gradients \cite{wang2016learning,duan2016rl}. This type of meta-learning is also appealing because it could learn a better inner-loop learning algorithm than any currently available machine learning algorithms. On simple reinforcement learning tasks it has been demonstrated that this approach can learn to explore, exploit, and balance between the two \cite{wang2016learning, duan2016rl}. It can also seemingly build an internal model of the world and plan within that model to behave like model-based RL methods instead of model-free RL methods, or at least the networks it produces pass (admittedly imperfect \cite{akam2015simple}) tests designed to tell model-based methods from model-free methods \cite{wang2016learning}.

Research on this pillar will involve improving the number and difficulty of tasks that can be solved. We must also identify what `materials' the learner (e.g.\ the RNN) should be made of. For example, the work in the second meta-learning camp mentioned so far was done with standard RNNs, where the only things that can change within the lifetime of the agent (i.e.\ during inner-loop training) are the activations of the network. The network weights do not change within its lifetime: they instead change only across `generations' (aka iterations) via the outer-loop optimization steps. We recently introduced an approach called \emph{differentiable plasticity} that enables meta-learning with a different type of neural network \cite{miconi2018differentiable}. The weights in these networks change intralife via Hebbian learning (in a fashion similar to the principle that  `neurons that fire together, wire together'). In this case, each weight consists of two parameters: one is fixed throughout the lifetime of the agent and is directly tuned by the outer-loop optimization algorithm. The other changes as a function of network dynamics, so the network has the ability to store information in its weights, which can improve performance on many tasks vs.\ traditional RNN meta-learning approaches \cite{miconi2018differentiable}. We showed that on some problems that require remembering lots of information over time, RNNs composed of these Hebbian materials outperform LSTMs composed of traditional neurons. While work on Hebbian learning in neural networks has a long history, our recent introduction of the ability to train such networks via gradient descent offers the possibility to harness this alternative to SGD at scale when meta-learning large, deep neural networks. 

Another type of learner network includes \emph{neuromodulation}, wherein the output of one neuron in a network can change the learning rates of other connections in the network \cite{ellefsen2015neural, velez2017diffusion,norouzzadeh2016neuromodulation,soltoggio2018born,risi2009novelty,soltoggio2012modulated}. As with Hebbian learning, there is a long history of neuromodulation work without using gradient descent, but we have recently introduced the ability to train networks with neuromodulation in a fully differentiable, end-to-end fashion \cite{miconi2018backpropamine}. An example type of problem that could in theory be solved with meta-learning and neuromodulation is catastrophic forgetting, a phenomenon wherein neural networks are unable to continuously learn because they learn each newly encountered task by overwriting their knowledge of how to solve all previously learned tasks \cite{french1992semi,mermillod2013stability,french1999catastrophic}. Neuromodulation can help because some neurons in the network can detect which task is currently being performed, and those neurons can turn learning on in the part of the network that performs that task and turn learning off everywhere else in the network. That can enable a neural network to learn new skills without cannibalizing old skills. 
We have shown that meta-learning with neuromodulation can create networks that perfectly solve catastrophic forgetting, meaning they have learned how to learn without forgetting, albeit in small networks and on simple problems \cite{ellefsen2015neural, velez2017diffusion}. Since the first version of this paper was published, we introduced a method called A Neuromodulated Meta-Learning algorithm (ANML), which meta-learns a solution to catastrophic forgetting for deep neural networks at the unprecedented scale of 600 sequential tasks (over 9000 SGD updates) \cite{beaulieu2020ANML}. 

These are just a few examples of how the type of learner used, including the materials it is made of, can have a meaningful effect on the quality and type of solution that results. The choice of materials is a hyperparameter of the meta-learning algorithm. Of course, they too could (and likely ultimately will) be searched for. 

Meta-learning is currently computational expensive. It requires training over a distribution of different tasks and often involves differentiating through the entire learning process. Computational limitations, memory limits, and optimization challenges are major hurdles for this work. They currently limit the complexity of tasks that we can train on, how long those tasks can last (e.g.\ how many time steps), and how many outer-loop (meta) optimization steps we can perform. Research to improve our ability to train such networks and do so on more and harder tasks will be critical to unlocking the power of meta-learning. 

One question raised by meta-learning learning algorithms is what the distribution of tasks should be for meta-learning. How narrow should it be? Should it widen over time? As has been pointed out, switching from normal learning to meta-learning changes the burden on the researcher from designing learning algorithms to designing environments \cite{gupta2018unsupervised}. But why stop there? I argue that we should instead learn the appropriate task distribution that learning agents should meta-learn on. That is AI-GA's Third Pillar, which I discuss next.

\subsection{The Third Pillar: Generating effective learning environments and training data}
\label{Pillar3learningEnvs}

It is unlikely that we will ever know how to directly program a generally intelligent agent. Instead, we will almost certainly have to train a learning agent on the right training data and/or in the right set of training environments. That raises the question as to what the correct training environments are. Even if we knew what the right set of training environments were, we likely would not get away with sampling from them randomly for training (that would be too inefficient, if it worked at all). Instead, we also have to know the right \emph{order} in which to present the tasks, which may itself be a function of the current learning capabilities of the learning agent. Just as human educational systems have developed curricula for how to teach human learners a variety of subjects, we likely will need to produce effective curricula in order to produce generally intelligent artificial learning agents. 

Manually identifying and creating each training environment required and their order would be an extremely challenging, slow, and expensive undertaking. It would be limited by our time, intelligence, and our ability to understand each AI student enough to design optimal learning environments for them. However, we know from the history of machine learning that data is the rocket fuel of success, and we will therefore need lots of it. 

The AI-GA philosophy is that we should have learning algorithms learn to automatically generate effective learning environments (i.e.\ generate their own training data), including creating an effective curriculum that can train up an initially unintelligent agent (the equivalent of a human newborn) into a generally intelligent agent (the equivalent of a human adult, or something smarter). Part of generating a learning environment is generating the reward function that defines success in that environment. One can imagine the generation of labeled training data (images and their labels), virtual worlds that focus on mastering a particular skill (much as a hockey player practices passing, shooting, and skating), and even entire virtual worlds in which agents learn to communicate, coordinate to perform complex tasks, and learn a variety of increasingly difficult concepts, from math and physics to writing, philosophy, and machine learning. One approach might be to create a set of different training environments and expose a learning agent to them in some order (possibly contingent on their past performance). Another approach might be to create a single training environment, which itself internally represents a curriculum for an agent. As on Earth, the difference might ultimately be semantic: in your lifetime have you experienced a set of different learning environments, or a single complex one? 

There has been little research into the Third Pillar. It is the least-studied, least-understood, and likely hardest of the three pillars. That also makes it an interesting new frontier of research. Many fundamental breakthroughs await in this nascent field, making it a ripe orchard in which researchers can hunt for impactful discoveries. I next outline a sketch of how we might make progress on this pillar, including some of our early work in this direction. 
Of course, the most exciting ideas and directions are the ones we cannot envision yet, making the following sketch more of an exercise to kickstart progress rather than a playbook for future research. 

Much work has occurred in the fields of computational evolutionary biology, artificial life, and neuroevolution attempting to investigate how to create an open-ended complexity explosion  like the one that occurred with living animals on Earth \cite{lenski2003evolutionary,stanley:open,bedau2000open,langton1997artificial,ray1993evolutionary,brant2017minimal,soros2014identifying,adami2000evolution,taylor2016open,huizinga2018evolving,kouvaris2015evolution,kounios2016resolving,stanley2019designing,clune2011selective,channon2019maximum}. The dominant theme is to try to create the conditions within a virtual world in which agents evolve and hope a complexity explosion will occur. In virtually all of that work, the environment itself is manually created by human researchers, including the rules by which agents interact if there are multiple agents. The hope is often that \emph{interactions} between such agents will kickoff a coevolutionary arms race that will produce open-ended innovation. 
A specific version of this idea, called coevolution, is to have two populations of learning agents competing or cooperating with each other
\cite{de2004incremental,ficici:alife98,jong2004ideal,popovici:hnc12,wiegand:gecco01}. A related idea called self-play is to have an agent learn by playing itself, or an archive of past versions of itself \cite{bansal2017emergent,silver2017mastering,Silver:2016aa}.
Significant gains have been made, most famously in learning to play the games Go and Dota \cite{silver2017mastering,Silver:2016aa}, validating both the idea of generating effective learning environments (i.e.\ generating training data) and having agents play against automatically learned curricula. However, these approaches have failed to create anything resembling an open-ended complexity explosion because the non-agent (abiotic) component of the environment they operate in is fixed. For example, while self-play has automatically produced agents that are capable of kicking soccer balls into a goal and playing the games of Go and Dota, no amount of self-play in virtual soccer, Go, or Dota will produce agents that can drive, write poetry, invent flying machines, and conduct science to understand the world they find themselves in. Another approach is to automatically generate different reward functions within an environment \cite{gupta2018unsupervised}. While promising, like self-play, ultimately such agents are confined to a predefined environment that limits what they can learn. 

In my opinion, a more promising direction is to explicitly optimize environments to be effective for learning, instead of hoping we can create environments that create dynamics that lead to coevolutionary arms races that produce generally intelligent learners. Just thinking about environments as something that can be optimized by learning algorithms is interesting and opens many new research directions. One project in this direction is PowerPlay \cite{schmidhuber2013powerplay,srivastava2013first}. Another vein of research comes from the community attempting to automatically generate content for video games, including training AI agents \cite{justesen2018illuminating,shaker2016procedural,khalifa2016general}. The first paper that made me consider directly optimizing environments as an interesting possibility is \citet{brant2017minimal}, where an expanding set of novel mazes were generated, each of which could be solved by at least one agent. We followed up on that idea with the Paired Open-Ended Trailblazer (POET) algorithm in \citet{wang2019paired}, which is discussed in more detail below. However, the question remains of how to properly conduct such optimization. What should the learning algorithm that is optimizing the environment try to make the environment accomplish? What is the reward function (aka loss function or fitness function) for the environment generator? This is one of the key questions for AI-GA research. 

One option is the following: define intelligence, then optimize an environment such that learners exposed to that environment are increasingly generally intelligent. This option seems unrealistic for a variety of reasons. The first two problems are that it requires defining what intelligence is and how to measure it. Those problems have eluded scientists and philosophers for centuries. The second is that, even if we did have a reward function for general intelligence, it would almost certainly be highly sparse and deceptive (non-convex). In the parlance of evolutionary biology, these would be considered fitness landscapes that are perfectly flat in many places and extremely rugged in others. The reward function would be sparse because for much of the space there would be no signal or gradient from the reward function. Stanley and Lehman provide a colorful example in the uselessness of, for example, trying to select for human-level intelligence by starting with bacteria and selecting for those that perform better on IQ tests \cite{stanley2015greatness}. Assuming for a second that IQ tests do have something to say about general intelligence, it is clearly unhelpful to try to guide a search for intelligence from bacteria by optimizing for performance on IQ tests because until you already have something with the intelligence of a human child, the test provides no signal to guide search. 

The problem of deception is even more pernicious. It describes the situation where the reward function provides not just zero signal, but the wrong signal, causing search algorithms to get trapped on local optima far lower in performance than the global optimum. When searching for the solutions to extremely ambitious problems, even if you have a way to quantifiably measure what success looks like, optimizing for that score is unlikely to produce the correct path through the search space to find the global optimum, and instead usually results in getting trapped on a local optima \cite{stanley2015greatness,lehman2011abandoning,woolley2011deleterious,nguyen2016understanding}. In other words, despite our best efforts, we will likely create deceptive reward functions that do not provide a gradient towards any satisfactory optimum, and will instead lead the search process to get trapped on low-performing local optima.  

All of these problems are why basic science is so important. If we went back a few hundred years and sought to create clean energy, it would have been futile to search for technologies that produce fewer and fewer carbon emissions per unit of energy produced. Instead, the first effective clean energy (nuclear power) was discovered by someone passionately thinking about light beams chasing them on a train moving at the speed of light. Similarly, to invent microwaves, one should not have rewarded engineers for making devices that are increasingly fast at heating food, but instead the correct path was to invest in radar technology.  If one looks at most major technological innovations, the path of stepping stones that led to them is nearly always circuitous and would have been unpredictable ahead of time \cite{stanley2015greatness}. 

Nevertheless, it is useful to recognize that it is possible to generate effective learning tasks with a particular task in mind, such as performing well on an IQ test, having a robot perform a specific skill, etc. While it is unlikely to be fruitful for very ambitious tasks (where the learner is far from possessing the desired skill), it could be practically useful if the amount to be learned is smaller. I call this approach to generating effective learning environments the \emph{target-task approach}. While it is helpful to research how to create learning algorithms that can produce learners that solve a predefined task of interest---a subject we are working on \cite{petroskisuch2019GTN}---for ambitious goals like producing general AI we likely will need a different approach. 

A second approach, which I believe is more likely to work for ambitious purposes like producing general AI, is to pursue open-ended search algorithms, meaning algorithms that endlessly generate new things \cite{stanley:open}. In the AI-GA context, that would mean algorithms that endlessly generate an expanding set of challenging environments and solutions to each of those challenges. The expectation is that eventually general AI will emerge as a solution to one or a set of the generated environments. I call this the \emph{open-ended approach} to creating effective learning environments. This path is what occurred on Earth. Natural evolution did not start out with the goal of producing general intelligence. Instead, a remarkably simple algorithm (Darwinian evolution) began producing solutions to relatively simple environments. The `solutions' to those environments were organisms that could survive in them. Those organism often created new niches (i.e.\ environments, or opportunities) that could be exploited. For example, the existence of gazelles creates a niche for cheetahs, which in turn creates a niche for microorganisms that live inside of cheetahs, etc. The existence of trees enables vines, epiphytes, and canopy butterflies, each of which creates niches for other organisms. The result is an ever-expanding, open-ended collection of new niches (challenges) and solutions to those niches (agents with a variety of skills). Ultimately, that process produced all of the engineering marvels on the planet, such as jaguars, hawks, and the human mind. Note that this approach avoids the need to create a definition and measurement of general AI. It allows us to rely on the fact that we will know it when we see it.

What happened in natural evolution was extremely computationally expensive. The AI-GA paradigm motivates research into the following question: can we create computational systems that \emph{efficiently} abstract the key ingredients that produce complexity explosions like those produced by evolution on earth? If so, we may be able to create algorithms that effectively endlessly innovate and could ultimately produce general AI. In other words, the goal is to create an AI-GA without requiring a planet-sized computer.  

We of course do not know all of the key ingredients that made natural evolution so effective, nor do we know the best way to abstract them. However, we have made progress in identifying some of these key ingredients and how to abstract them. I will attempt to briefly summarize the work I consider promising in this direction. 

\subsubsection{Encouraging behavioral diversity}

An essential ingredient in the complexity explosion that occurred on earth is the creation of a diverse set of niches and organisms that can live in those niches. An abstract property that will almost certainly be key is thus diversity. For decades, researchers have known that diversity can help search algorithms get off of local optima. But the vast majority of approaches have encouraged diversity in the original search space, such as the weights of a neural network that controls a robot. Such search spaces are high-dimensional and often diversity in the weights of neural networks is not \emph{interesting} diversity. For example, there are a near infinite number of weight vectors that can make a robot immediately fall over. A key insight came when \citet{lehman2011abandoning} emphasized the importance of encouraging \emph{behavioral diversity}. The idea is to define a behavior space in which diversity is interesting, and then search for neural network weight vectors that produce different behaviors in that space. For example, imagine a simulation of the city of San Francisco. We might want to encourage a search algorithm to produce robots that visit not-yet-visited $x,y,z$ locations in San Francisco. Visiting new $x,y,z$ locations would require learning skills like walking, avoiding obstacles, climbing stairs, taking elevators, opening doors, crossing streets, climbing trees, picking locks, etc. In fact, \citet{lehman2011abandoning} showed that searching for novelty \emph{only} (and completely ignoring the reward function), a algorithm they call \emph{Novelty Search}, can solve problems that are not solvable with objective-driven search when local optima are present. We subsequently showed that novelty search does in fact learn general skills for how to move about an environment that can transfer to new environments \cite{velez2014novelty}. In an algorithm called \emph{Curiosity Search} \cite{stanton2016curiosity}, we later showed that encouraging an agent to visit as many different locations within its lifetime (i.e.\ within an episode) as possible improved performance on the Atari game Montezuma's Revenge, a notoriously difficult, sparse, hard-exploration problem, matching the then state of the art \cite{stanton2018deepCS}. There have been many other important recent papers investigating how much agents can learn when motivated by curiosity alone \cite{burda2018large,pathak2017curiosity,oudeyer2009intrinsic,savinov2018episodic,eysenbach2018diversity,gupta2018unsupervised,oudeyer2007intrinsic,schmidhuber2010formal}.

While fascinating, rewarding novelty alone is unlikely to reproduce the complexity explosion that occurred on Earth. One missing ingredient is that on Earth, while the creation of niches creates new environments (which indirectly create a pressure to produce diverse behaviors), within each niche there is a pressure to be high-performing. Such intra-niche competition is the most common way people think of `survival of the fittest', wherein faster cheetahs replace slower ones, stronger elephant seals outcompete weaker rivals for control of the harem, etc. The most straightforward way to combine a pressure for novel behaviors and high performance is to optimize for both of them. Many have shown such a combination is effective, whether in a weighted combination of novelty and reward \cite{conti2018improving}, multi-objective algorithms \cite{Mouret2012,mouret2009overcoming}, or algorithms that intelligently rebalance between novelty and performance as a function of learning progress \cite{conti2018improving}. However, while the novelty pressure helps avoid local optima, such algorithms still tend to produce one or a few variations on the single theme that search converges on. They do not produce a wide variety of different, yet high-quality solutions. 

\subsubsection{Quality Diversity algorithms}

Ultimately, what we want are algorithms that create a diverse set of solutions where each of those solutions is as high-performing as possible for that type. For that reason, my colleagues and I have created a new family of search algorithms called \emph{Quality Diversity (QD) algorithms}. The first of its kind was \emph{Novelty Search with Local Competition}   \cite{lehman2011evolving} followed by MAP-Elites \cite{mouret2015illuminating}. The general idea is to define (or learn) a low-dimensional behavioral (or phenotypic) space and then explicitly search for the highest-performing solution in each region of that space. For example, if one was searching in the space of weights of a deep neural network that encodes the morphology of a robot body, the performance criterion might be walking speed and the space we want to encourage diversity in could be the height and weight of the robot. The QD algorithm would thus search for and return the fastest tall, skinny robot, the fastest tall, fat robot, the fasted short, skinny robot, etc. 

QD algorithms have already been harnessed to solve very difficult machine learning problems. For example, in a paper in Nature \cite{cully2015robots}, we harnessed QD to obtain state-of-the-art robot damage recovery, wherein a robot can adapt to substantial damage in 1-2 minutes. We also showed that with access to a deterministic simulator for training, an enhanced version MAP-Elites called Go-Explore can solve the hard-exploration benchmark challenges of the Atari games Montezuma's Revenge and Pitfall, dramatically improving the state of the art \cite{ecoffet2019go}.  

One essential component of QD algorithms that fuels their success is \emph{goal switching} \cite{nguyen2016understanding}. The idea is that within a niche a parameter vector is optimized to perform well, but if at any point it turns out that a perturbed version of that parameter vector belongs to another niche and is better than its current champion, that perturbed parameter vector becomes the elite in the other niche. That enables one style of solutions in a niche (which may be stuck on a local optima) to be disrupted by another type of solution that was discovered via optimization on a different niche. This captures a dynamic similar to scientific and technological innovation (e.g.\ how microwaves invaded the kitchen niche despite their underlying technology being developed for radar, a very different purpose). It also captures the dynamic of natural evolution where a species optimized for one niche can invade a new niche if it is superior.
Research has shown that such goal switching significantly improves the performance of QD algorithms \cite{nguyen2016understanding, mouret2015illuminating, huizinga2018evolving}. For example, we showed how it can enable the creation and combination of different skills to produce multi-modal robots capable of jumping, crouching, running, and turning \cite{huizinga2018evolving}. 

QD algorithms can be used in many types of problem domains, such as solving different types of mathematical, writing, or musical challenges. For example, in our work on \emph{Innovation Engines} \cite{nguyen2016understanding}, we showed that the same ideas underlying QD algorithms could be used to generate different types of high-quality, recognizable images. Goal-switching proved essential to avoid local optima and was a catalyst for innovation. We also observed in that work another hallmark of biological complexity explosions: adaptive radiation. Innovations in one niche rapidly spread to many other niches, allowing key innovations to spread and become the foundation upon which additional innovations specific to different niches are built. This phenomenon is reminiscent of the Cambrian Explosion, wherein the innovation of the four-legged body plan led to a rapid radiation of that motif into a large number of different niches to ultimately produce all of the diverse four-legged creatures on Earth. 

Another benefit of the goal-switching in QD algorithms is the creation of better underlying representations for search. If a structure is repeatedly optimized (or selected) to move in some dimensions and not others, it can restructure itself to make traversing the preferred dimensions of variation more likely than traversing non-preferred dimensions. In the literature of biological and computational evolution, this phenomenon is called \emph{evolvability} \cite{kounios2016resolving,kouvaris2015evolution,wagner1996perspective,conrad1979bootstrapping,pigliucci2008evolvability,wagner2013robustness,mengistu2016evolvability}. \citet{lehman2011abandoning} showed that Novelty Search produces more evolvability than objective-based search. The reason is intuitive. With objective-based search, the optimization algorithm myopically adds any hack to the representation that improves performance. Like technological debt in software engineering, adding hacks upon hacks without paying the cost of refactoring (which temporarily makes things worse before they ultimately get much better) leads to code that is difficult to adapt to new use cases. In contrast, Novelty Search encourages the production of new behaviors, and thus perturbations that make major changes are less likely to be immediately rejected \cite{lehman2010efficiently}. Specifically, Lehman and Stanley showed that neural networks subjected to novelty search were more compact than objective-based search \cite{lehman2011abandoning,lehman2011improving}, in favor of the hypothesis that they are more evolvable. Novelty search was also shown to lead to representations that produce more behavioral diversity when perturbed, a measure of evolvability \cite{lehman2011improving}. However, we found they were not faster in adapting to a new tasks \cite{velez2014novelty}, which is contrary to this hypothesis, although more research needs to be done on this question.  Additionally, we discovered that when something similar to a neural network encodes pictures and these networks are subject to constantly changing, human-defined goals, the resulting representations are significantly smaller (a proxy for evolvability), more modular, and more hierarchical \cite{huizinga2017emergence}. Moreover, we found they are much more likely to produce sensible variations than nonsensical ones, and do so in a hierarchical way. For example, an image of a face might be changed to enlarge or shrink the entire face, or both eyes, or just one eye. Or a smile could be easily converted into a frown, etc. Less likely were changes that transformed a face into an eagle or a scrambled mess of pixels. When automatically generating images with Innovation Engines, we also found that goal-switching led to significantly more compact, adaptable representations \cite{nguyen2016understanding}.

\subsubsection{Environment-generating quality-diversity algorithms}

QD algorithms as originally conceived could create a high-quality set of diverse behaviors within a single, pre-defined environment only. However, for AI-GA's Third Pillar we need to generate different types of environments and their solutions. That desire motived our recent Paired Open-Ended Trailblazer (POET) algorithm \cite{wang2019paired}. In it, a parameter vector $\theta_{E}$ specifies an environment. In our demonstration domain, the environments were obstacle courses that could have different degrees of being hilly, gaps in the ground of varying width, and tree trunks of varying height. An additional parameter vector $\theta_{A}$ contains the weights of a DNN that controls an agent, which in this case is a robot that has to traverse the obstacle course as quickly as possible. An initial agent $\theta_{A^1}$ is optimized to solve the initial, simple environment $\theta_{E^1}$. Once the performance of $\theta_{A^1}$ is good enough, we copy $\theta_{E^1}$ and change it slightly to create $\theta_{E^2}$, which now represents a different environment.
A copy of $\theta_{A^1}$ is made to create $\theta_{A^2}$, which is then optimized to solve the new environment $\theta_{E^2}$. Importantly, optimization of $\theta_{A^1}$ on $\theta_{E^1}$ continues in parallel. Over time, environments are periodically generated by copying any of the current environments for which its agent performs above some threshold. All or a subset of the current agents are evaluated on each new environment. Environments are kept only if they are not too hard for all of the agents in the population, or are not too easy for any of the agents. A copy of the highest-performing agent is transferred to the new environment, where it begins optimizing to try solve it. 

POET exhibits many desirable properties. First, it creates an expanding set of diverse environmental challenges, each of which can be solved by the current population of agents to some degree. In most cases, agents get better at solving their particular challenge, meaning they are gaining skills. Additionally, agents can goal-switch between challenges. We observe that in many cases the current agent in an environment is stuck on a local optima, but eventually an agent from another environment transfers in, ultimately leading to much higher performance. One can imagine that POET could create wildly different species of solutions within one run, but our initial demonstration domain was too simple to produce tremendous diversity because the environmental encoding only allowed the production of environments with different amounts of landscape ruggedness and different sizes of gaps and obstacles. However, if POET was combined with a flexible way to encode environments, one of its runs could create water worlds, desert worlds, and mountain worlds, each with its own types of agents customized to perform well in those worlds. Such specialization is especially easy to imagine if the bodies of virtual robots are simultaneously optimized, which is a subfield with a long history \cite{sims1994evolving,cheney2013unshackling,cheney:alife14,hornby2002creating,auerbach:bodybrain,ha2018evolveBodies}. Presumably goal-switching would happen much more often within certain types of water worlds, less so amongst more different types of water worlds, and never between water worlds and mountain worlds (at least, not directly). Thus, such algorithms naturally would hedge their bets on the best path to creating any type of interesting solution, including general AI, by simultaneously pursuing a diversity of high-quality solutions. In effect, POET creates multiple, simultaneous, overlapping curricula to learn an ever-expanding set of skills. Many of the curricula might be ineffective dead ends, but as long as some of them are fruitful, the algorithm can succeed. 

\subsubsection{More expressive environment encodings}

One drawback to the POET work is that it assumes a specific type of world, such as obstacle courses in a particular physics simulator, and a specific way of parameterizing that type of world (e.g.\ a vector with numbers defining the width of the gaps, the height of the stumps, etc.). But ultimately such a strategy is confined to only be able to produce environmental challenges allowable by that parameterization of that physics simulator. For example, our obstacle course simulator does not allow the creation of many types of challenges, such as doors, tire swings, swimming worlds, playing chess, or needing to learn different types of mathematics. It also cannot generate sound, smell, and other sensory modalities. In short, the original POET did not have a sufficiently expressive environmental encoding. 

In our newest work on the Third Pillar \cite{petroskisuch2019GTN}, we are creating algorithms that can generate effective learning environments with a \emph{fully expressive environment encoding}, meaning one that can generate all possible learning environments. Recalling the concept of computers that are Turing Complete, we might call an environmental encoding that can create \emph{any possible learning environment} Darwin Complete. The name reflects that the encoding enables the creation of all of the environments that made Darwinian evolution successful (and many more). 

To create a Darwin Complete environment encoding, we generate environments via a deep neural network (that can optionally be recurrent). We call these DNNs \emph{Generative Teaching Networks} \cite{petroskisuch2019GTN} because we explicitly train them to be optimal teachers for another student deep neural network that learns on the data (or in the environment) the GTNs create. Because recurrent neural networks are Turing Complete \cite{siegelmann1995computational}), GTNs can produce any type of data, and thus could in theory create everything from image classification tasks to entire virtual 3D worlds complete with sound, touch, and smell. They could also create opponent agents to play against, and thus are strictly more expressive than self-play techniques (although they are likely much harder to optimize). That said, it may be easier to have trained agents interact within GTN-produced worlds, rather than forcing the GTN to learn to produce the policies of an arbitrary number of agents.
In that option, there would be an entire population of agents in each generated environment. That would potentially make it easier to create multi-agent interactions, including the emergence of language and culture (including the catalyzing force of agents learning via cultural transmission and the cultural ratchet, meaning the amassing of increasing amounts of knowledge over time that each agent can learn from \cite{tomasello2009cultural}).

Because DNNs are differentiable, we can have GTNs produce training data or environments for a learner DNN, then test the trained learner on a target task, and differentiate back through the entire learning process to update the parameters of the GTN to improve its ability to produce effective training data. Because the GTN is trying to make the learner good on a predefined task, this use case is an example of the ``target-task'' version of generating effective learning environments. In our experiments so far, we have shown that a GTN can be trained to produce data that enables a student network to learn to classify MNIST. Interestingly, the GTN is not constrained to produce realistic-looking training data (e.g.\ images a human would recognize as handwritten digits). In fact, we found that the data it generates look completely unrecognizable and alien, yet still the student network learns to recognize real handwritten digits. Moreover, the student network learns four times faster than when training on real MNIST training data! The result that unrecognizable images are meaningful to DNNs is reminiscent of the realization that deep neural networks are easily fooled and will declare that unrecognizable images are everyday objects (e.g.\ guitars and starfish) with near certainty \cite{nguyen2015deep}. 
It is an interesting, open question as to whether natural brains, including those of humans, could be rapidly trained to perform any skill via such alien data (a la the novel Snow Crash). Interestingly, researchers recently showed that they could generate fooling images (`super-stimuli,' in the parlance of biologists) for the neurons of real monkeys \cite{ponce2019cell}; they did so by generating images via the DGN-AM technique \cite{nguyen2016synthesizing} to synthetically generate data that activate neurons in a live monkey's brain. Many of the synthesized images were unrecognizable, yet activated the neurons in the brains of monkeys more than any of real images from the natural world. 

As mentioned above, there are two options for generating effective learning environments. The previous paragraph describes how we have already experimented with GTNs in the target-task paradigm. Intriguingly, GTNs (or any sufficiently expressive environment generator) can also be used in the open-ended paradigm for generating effective learning environments. This is a promising path to making progress on AI-GA's Third Pillar. The idea is to harness GTNs to produce an expanding set of learning challenges for agents. For example, one could pair the GTN encoding with POET to create an expanding set of GTNs that each specify an environment. Alternately, a single, powerful GTN could be created that is conditioned on a noise vector (and possibly an agent's past experience and learning progress) to endlessly produce new, effective learning challenges. Of course, just because the GTN can do that in theory does not mean it will be easy to make it work in practice, and much work remains to accomplish that lofty goal. Additionally, GTNs are just one approach to generating learning environments. They may prove \emph{too} expressive and thus make searching their vast search space intractable. We might want to bake in more prior knowledge by constraining all environments to be in a physics simulator (based on our laws of physics), which both narrows the search space considerably and increases the chance that the skills learned will be relevant to our world and more comprehensible to us. Many other approaches are viable and research into the best ways to generate effective learning environments, including how to encode them, will be essential for progress. 

A wide open question in this line of research is what the reward function for the environment generator should be in the open-ended version of generating effective learning environments. This is a deep, fascinating, hard question that could be a key to unlocking significant progress in machine learning research. Finding the answer to this question could finally enable us to solve the longstanding grand challenge of producing open-ended search algorithms \cite{stanley:open} and producing general AI, potentially solving two grand challenges in one stroke. 

I do not have an answer to what this environment-generator reward function should be, but I have some ideas that researchers could begin experimenting with and improving on. The question relates to abstracting \emph{what environments were for} in natural evolution or, analogously, what \emph{problems were for} in the history of scientific and technological innovation. What role did environments serve in producing the complexity explosion on Earth, including creating human intelligence? What role did problems play in scientific and technological innovation within our culture? 

What I consider the most promising direction for potential reward functions for environmental generators is to define environments as useful (and thus reward their creation) if the environments are such that agents transferring in from other, previously generated environments (1) perform well in the new environments (after some learning) faster than if they were trained on them from scratch, and (2) learn in the new environments (i.e.\ the environments are not too easy and not too hard, such that the agent experiences `learning progress' \cite{schmidhuber1991curious}). The first condition encourages shared structure between the problems (e.g.\ similar laws of physics, or mathematical rules), such that having learned in some subset of problems makes agents be able to transfer that knowledge to other environments. That could prevent the creation of arbitrarily different problems that are challenging, but in uninteresting ways. The second condition prevents the creation of worlds that do not require and encourage learning. It also forces environments to be new in some way. This idea needs to be developed, improved, and experimentally investigated, but it gives a hint of how we can begin to make progress on producing very general principles for preferring the creation of some environments versus others. Another idea that could help is an explicit pressure to produce generalists, perhaps by incentivizing agents to be able to solve as many niches as possible. We have other ideas for how to reward the generation of interesting environments, but I cannot share them because we are actively investigating them. It is likely that none of these ideas will just work. Instead, we need lots of research by the community into these questions to make progress on this front. 

A major open question that remains is how we can constrain the generation of environments to be those we find interesting and/or that produce intelligence that helps us solve real-world problems. In other words, how do we ground the environment generator to make things relevant to us and our universe? For example, one might argue that such a system could produce intelligence that is alien to us and that we cannot communicate with. However, if it is truly general intelligence, presumably through its learning efforts and our own we could learn to communicate with it. Additionally, creating alien forms of intelligence would be fascinating as it would teach us about the limits and possibilities within the space of intelligent beings. 

\subsubsection{The viability of the Third Pillar}

The hypothesis behind the Third Pillar is that coming up with general principles for how to create effective learning environments is easier than creating a curriculum of training environments by hand. As with the overall AI-GA philosophy, the bet is that with sufficient compute, letting learning algorithms solve the problem will ultimately be easier than trying to solve it ourselves. Of course, this may require orders of magnitude more compute than we have at present, and whether we will have sufficient compute to see this pillar succeed before general AI is produced by the manual path is an open question. However, I predict that we will abstract the role environments play in the creation of complexity and intelligence, and that doing so will shave orders of magnitude off the total amount of compute required, such that we will not need the computation that was required on Earth to produce humans in order to create general AI via an AI-GA. 

One thing in favor of the Third Pillar, and the AI-GA approach in general, is that we know it can work. Darwinian evolution on Earth is the only existence proof we have of how to build general intelligence, and AI-GAs are modeled off of that phenomenon. It is a fascinating research grand-challenge to figure out whether we can extract the principles that made Earth's intelligence-generating algorithm so successful in a way that enables us to create an AI-GA with the computation available to us. 

Note that the AI-GA grand challenge (creating an algorithm that generates general AI) is related to, but different from, the grand challenge of creating open-ended search algorithms \cite{stanley:open}. One could create an AI-GA without it being open-ended. For example, one could launch an AI-GA with the target task being to pass the Turing Test, and the algorithm would then halt when it produces general AI because it would have no other goal to optimize for. One could also create an open-ended algorithm that would never produce an AI-GA (e.g.\ one that generates endless music innovations). That said, these two grand challenges are deeply related and will catalyze the progress of each other. 

\section{Discussion}
\label{discussion}

The AI-GA approach raises many questions. This section attempts to quickly address a variety of those different questions. 

Traditional machine learning involves hand-designing an environment, hand-selecting an architecture, and then hand-designing a learning algorithm that learns to solve that environment. Meta-learning represents a step towards a more fully automated pipeline. One area of research in meta-learning focuses on learning architectures. Another area of meta-learning is learning the learning algorithm. But in both, a researcher is still required to design the distribution of tasks for meta-training. AI-GAs take the next step in this natural progression by automatically learning all three things: the architecture, the learning algorithm, and the training environments. 

That said, the AI-GA approach is not free of building blocks. It does not start from scratch, trying to create intelligence from nothing but the laws of physics and a soup of atoms. It might, for example, start with the assumption that we want neural networks with regular and neuromodulatory neurons, niches and explicit goal-switching, and populations of agents within each environmental niche. Research on AI-GAs will involve identifying the minimum number of sufficient and catalyzing building blocks to create an AI-GA. But the set of building blocks AI-GA researchers will look for will be very different from those for the manual engineering approach. For example, AI-GA will likely attempt to automatically learn everything in Table \ref{tableOfBBs}. Rather than discovering those building blocks manually, AI-GA researchers will try to figure out what are the building blocks that enable us to abstract open-ended complexity explosions in a computationally tractable way. I hypothesize that fewer building blocks need to be identified to produce an AI-GA than to produce intelligence via the manual AI approach. For that reason, I further hypothesize that identifying them and how to combine them will be easier, and thus that AI-GA is more likely to produce general AI faster. That said, an open question is how much compute is required to make an AI-GA. If we magically had vastly more compute available starting tomorrow, my confidence in AI-GAs producing general AI before the manual path would be greatly increased. What is less clear is whether AI-GAs can beat the manual path given that the computation AI-GAs need is not yet available. The lack of computation gives the manual path an advantage since it is more compute-efficient. On the other hand, as Rich Sutton has argued, history often favors algorithms that better take advantage of the exponential increases in computation the future provides \cite{sutton2019bitter}. 

It may turn out that the manual path can succeed without having to identify hundreds of building blocks. It might turn out instead that only a few are required. At various points in the history of physics many pieces of theoretical machinery were thought to be required, only later to be unified into smaller, simpler, more elegant frameworks. Such unification could also occur within the manual AI path. But at some point if everything is being learned from a few basic principles, it seems more like an AI-GA approach instead of a manual identify-then-combine engineering approach. In other words, the AI-GA approach \emph{is} the quest for a set of simple building blocks that can be combined to produce general AI, so true unification would validate it. 

\new{
One interesting way to catalyze the Third Pillar is to harness data from the natural world when creating effective learning environments. This idea could accelerate progress in both the target-task and/or open-ended versions of the Third Pillar. For example, the environment generator could be given access to the Internet. It could then learn to generate tasks that involve classifying real images, imitating the skills animals and/or humans perform (e.g. in online video repositories such as Youtube), solving problems in existing textbooks, or solving existing machine learning benchmarks in language, logic, reinforcement learning, etc. There is a long history of fruitful research in imitation learning and learning via observation that demonstrates the benefits of exploiting such data \cite{edwards2019imitating,behbahani2018learning,stadie2017third,aytar2018playing,salimans2018learning,ecoffet2019go,torabi2018behavioral,pomerleau1989alvinn,ng2000algorithms,abbeel2004apprenticeship}. AI-GAs too could benefit from this treasure trove of information. Incorporating such tasks might also have the benefit of making the AI that results more capable at solving problems in our world and better able to communicate with and understand us (and vice versa). }

There is a third path to general AI. I call it the ``mimic path.'' It involves neuroscientists studying animal brains, especially the human brain, in an attempt to reverse engineer how intelligence works in animals and rebuild it computationally. In contrast to abstracting the general principles found via neuroscience and building them in any workable instantiation (which is the purview of the manual path), the mimic path attempts to recreate animal brains in as much faithful detail as possible. The most famous example of this approach is the work of  Henry Markram and his Blue Brain and European Human Brain projects. This path is also independently worthwhile, and should be pursued irrespective of whether the other two paths have already succeed. That is because if the manual or AI-GA paths build general AI that does not resemble human brains, we would still be interested in specifically how human intelligence works. If the mimic path wins, we would still be interested in the other two paths for the reasons outlined earlier. The mimic path is unlikely to be the fastest path to producing general AI because it does not seek to benefit from the efficiencies of abstraction. For example, if an entire neocortical column could be functionally replaced by a multi-layer recurrent neural network with skip connections, the spirit of the mimic path would be to eschew that option in favor of faithfully replicating the actual human neocortical column with all of its expensive-to-simulate complexity. Additionally, the mimic path is slowed by the difficulty of creating the technologies required to identify what is happening inside functioning natural brains. Overall, I do not view the mimic path it as one of the major paths to producing general AI because the goal of those committed to it is not solely to produce general AI. I thus only mention it this late in this essay, and still consider the manual and AI-GA paths as the two main paths. 

For the most part, this article has assumed that neural networks have a large part to play in the creation of AI-GAs. That reflects my optimism regarding deep neural networks, as well as my experience and passions. That said, it is of course possible that other techniques may produce general AI. One could substitute other techniques into the pillars of the AI-GA framework. For example, one could replace Pillar One and Pillar Two with Solomonoff induction via AIXI \cite{hutter2004universal,hutter2000theory} or similar ideas. To do so, major advances would be required to make these approaches computationally tractable, as with DNNs. However, one would still need work on AI-GA's Third Pillar because Solomonoff induction and AIXI take as a starting assumption a single environment to be solved. That highlights again that generating effective learning environments is the newest, least-explored area of research of the three pillars, even in far disparate areas of AI research. Advances in it may thus benefit many different subfields of AI. 

\new{
I also want to emphasize that AI-GAs are \emph{not} an ``evolutionary approach,'' despite people gravitating towards calling it that despite my saying otherwise. There are a few reasons to avoid that terminology. A first reason is that many methods could serve as the outer-loop optimizer. For example, one could use gradient-based meta-learning via meta-gradients \cite{petroskisuch2019GTN,finn2017model,maclaurin2015gradient} or policy gradients \cite{wang2016learning,duan2016rl}. Additionally, one could potentially use Bayesian Optimization as the outer-loop search algorithm, although innovations would be needed to help them search in high-dimensional search landscapes. 
Of course, evolutionary algorithms are also an option. There are pros and cons to using evolutionary methods \cite{such2017deep,lehman2018more,salimans2017evolution,wierstra2014natural,stanley2019designing}, and benefits to hybrid approaches that combine evolutionary concepts (searching in the parameter space) with policy-gradient concepts (searching in the action space) \cite{plappert2017parameter,fortunato2017noisy}. Because they are just one of many choices, calling AI-GAs an evolutionary approach is inaccurate. 
A second reason to avoid calling this an evolutionary approach is that many people in the machine learning community seem to have concluded that evolutionary methods are not worthwhile and not worth considering. There is thus a risk that if the AI-GA idea is associated with evolution it will not be evaluated with a clear, objective, open-mind, but instead will be written off for reasons that do not relate to its merits. 
}

Of course, in practice the three different paths will not exist isolated from each other. The manual, mimic, and AI-GA paths will all inform each other as discoveries in one catalyze and inspire work in the other two. Additionally, people will pursue hybrid approaches (e.g. the vision outlined in \citet{botvinick2017building}) and it will be difficult in many cases to tell which path a particular group or project belongs to. That said, it is instructive to give these different approaches their own names and talk about them separately despite the inevitable cross pollination and hybridization that will occur. 

\new{AI-GAs are a bet on learning in simulated worlds. The hypothesis is that we could create general AI entirely in such virtual worlds (although they may have to be complex worlds). However, once \emph{general} AI is produced, that AI would be a sample-efficient learner that could efficiently learn in our world. For example, such an agent could be transferred to a robot that could learn to maneuver in our world, including helping and otherwise interacting with humans. Thus, rather than specifically designing techniques to transfer from simulation to reality, the approach would not require such adaptation because the general AI would perform that efficiently itself.}

Many researchers will enjoy research into creating AI-GAs. There are many advantages to AI-GA research versus the manual path. Initially, there are currently tens of thousands of researchers pursuing the manual path. There are very few scientists currently pursuing AI-GAs. For many, that will make AI-GA research more exciting. It also decreases the risk of getting scooped. Additionally, to pursue the manual path, one might feel (as I have historically) the need to stay abreast of developments on \emph{every} building block one considers potentially important to creating a large, complicated general AI machine. One might thus feel a desire to read all new, important papers on each of the building blocks in Table \ref{tableOfBBs}, as well as building blocks it does not list and new ones as they are discovered. That is of course impossible given the number of such papers published each year. Because AI-GAs have fewer researchers and, if my hypothesis is correct, fewer building blocks required to make them, staying abreast of the AI-GA literature should prove more manageable. Additionally, if the large-scale trend in machine learning continues, wherein hand-coded components are replaced by learning, working on AI-GAs is a way to future-proof one's research. To put it brashly, knowing what you  know now, if you could go back 15 years ago, would you want to dedicate your career to improving HOG and SIFT? Or would you rather invest in the learning-based approaches that ultimately would render HOG and SIFT unnecessary, and prove to be far more general and powerful? 

As AI-GA research advances, it will likely generate many innovations that can be ported to the manual path. It will also create techniques and narrow AIs of economic importance that will benefit society long before the ambitious goal of producing general AI is accomplished. For example, creating algorithms that automatically search for architectures, learning algorithms, and training environments that solve tasks will be greatly useful, even for tasks far less ambitious than creating general AI. Thus, even if creating an AI-GA ultimately proves impossible, research into it will be valuable. 

There has been a long debate in the artificial intelligence and machine learning community as to where we should be on the spectrum between learning everything from scratch and hand-designing everything. Some argue that we need to inject lots of human priors into a system in order for it to be sample-efficient and/or generalize well. Others think we should learn as much as possible from data, eschewing human priors because learned solutions are ultimately better once there is sufficient compute and data. Where does the AI-GA paradigm fit in this debate? First, where we should be on the spectrum of course depends on our goals and how much time we have to achieve them. If we already know how to build a solution that performs well enough and performance is all we care about, we should clearly do that. Additionally, if we want to build a solution on a short timeframe, it is likely that the fastest path to a decent solution will involve lots of human prior information (unless we already know how to create learning-based solutions, as is the case in the many areas where deep learning is currently the dominant technique). However, for more ambitious goals over longer time horizons, it is often the case that betting on a learning-based solution is a good strategy. 

However, a commitment to learning is not a commitment to sample inefficient learners that only perform well with extreme levels of data and that generalize poorly. 
The AI-GA philosophy is that via a compute-intensive, sample-inefficient outer loop optimization process we can produce learning agents that are extremely sample efficient and that generalize well. Just as evolution (a slow, compute-intensive, sample-inefficient process) produced human intelligence, as AI-GAs advance they will produce individual learners that increasingly approach human learners in sample efficiency and generalization abilities. One might argue that means that the system itself is not sample efficient, because it requires so much data. That is true in some sense, but not true in other important ways. One important meaning of sample efficiency is how many samples are needed \emph{on a new problem}. For example, a new disease might appear on Earth and we may want doctors or AI to be able to identify it and make predictions about it from very few labeled examples.  If an AI-GA produces anything akin to human-level intelligence, the learner produced would be able, as humans are, to be sample efficient with respect to this new problem. The idea behind AI-GAs is that as compute becomes cheaper and our ability to generate sufficiently complex learning environments grows, we can afford to be sample inefficient in the outer loop in the service of producing a learner that is at or beyond human intelligence in being sample efficient when deployed on problems we care about. Putting aside its computational cost, AI-GAs thus in some sense represent the best of both ends of the spectrum: it can learn from data and not be constrained by human priors, but can produce something that, like humans themselves, contain powerful priors and ways to efficiently update them to rapidly solve problems. 

\section{Safety and ethical considerations}
\label{ethics}

Any discussion of producing general AI raises the ethical question of whether we should be pursuing this goal. The question of whether and why we should create general AI is a complicated one and is the focus of many articles \cite{russell2015research,everitt2018agi,amodei2016concrete,brundage2016artificial,bostrom2014ethics}. I will not delve into that issue here as it is better served when it is the sole issue of focus. However, the AI-GA path introduces its own unique set of ethical issues that I do want to mention here. 

In my view, the largest ethical concern unique to the AI-GA path is that it is, by definition, attempting to create a runaway process that leads to the creation of intelligence superior to our own. Many AI researchers have stated that they do not believe that AI will suddenly appear, but instead that progress will be predictable and slow. However, it is possible in the AI-GA approach that at some point a set of key building blocks will be put together and paired with sufficient computation. It could be the case that the same amount of computation had previously been insufficient to do much of interest, yet suddenly the combination of such building blocks finally unleashes an open-ended process. I consider it unlikely to happen any time soon, and I also think there will be signs of much progress before such a moment. That said, I also think it is possible that a large step-change occurs such that prior to it we did not think that an AI-GA was in sight. Thus, the stories of science fiction of a scientist starting an experiment, going to sleep, and awakening to discover they have created sentient life are far more conceivable in the AI-GA research paradigm than in the manual path. As mentioned above, no amount of compute on training a computer to recognize images, play Go, or generate text will suddenly become sentient. However, an AI-GA research project with the right ingredients might, and the first scientist to create an AI-GA may not know they have finally stumbled upon the key ingredients until afterwards. That makes AI-GA research more dangerous. 

Relatedly, a major concern with the AI-GA path is that the values of an AI produced by the system are less likely to be aligned with our own. One has less control when one is creating AI-GAs than when one is manually building an AI machine piece by piece. Worse, one can imagine that some ways of configuring AI-GAs (i.e.\ ways of incentivizing progress) that would make AI-GAs more likely to succeed in producing general AI also make their value systems more dangerous. For example, some researchers might try to replicate a basic principle of Darwinian evolution: that it is `red in tooth and claw.' If a researcher tried to catalyze the creation of an AI-GA by creating conditions similar to those on Earth, the results might be similar. We might thus produce an AI with human vices, such as violence, hatred, jealousy, deception, cunning, or worse, simply because those attributes make an AI more likely to survive and succeed in a particular type of competitive simulated world. Note that one might create such an unsavory AI unintentionally by not realizing that the incentive structure they defined encourages such behavior. In fact, it is routine for researchers to be surprised by the strategies AI comes up with to optimize the objectives it is given, and often the strategies and behaviors it creates are not at all what the researcher intended \cite{lehman2018surprising}. That phenomenon will only increase as AI-GA systems become more powerful, complex, and are able to make significant advances on their own. If a system based on red-in-tooth-and-claw competition is a faster path to a successful AI-GA, or even if it is thought to be, then some organizations or individuals on Earth may be incentivized to create such systems despite their potential risks.

Additionally, it is likely safer to create AI when one knows how to make it piece by piece. To paraphrase Feynman again, one better understands something when one can build it. Via the manual approach, we would likely understand relatively more about what the system is learning in each module and why. The AI-GA system is more likely to produce a very large black box that will be difficult to understand. That said, even current neural networks, which are tiny and simple compared to those that will likely be required for AGI, are inscrutable black boxes that are very difficult to understand the inner workings of \cite{nguyen2015deep,nguyen2016multifaceted,nguyen2017plug,nguyen2016synthesizing,yosinski2015understanding,li2016convergent, yosinski2014transferable,szegedy2013intriguing,mahendran2014understanding,zeiler2014visualizing,li2015visualizing,greydanus2017visualizing}. Once these networks are larger and have more complex, interacting pieces, the result might be sufficiently inscrutable that it does not end up mattering whether the inscrutability is even higher with AI-GAs. While ultimately we likely will learn much about how these complex brains work, that might take many years. From the AI safety perspective, however, what is likely most critical is our ability to understand the AI we are creating right around the time that we are finally producing very powerful AI.  

For all these reasons, it is essential to invest in AI-GA-specific AI safety research. AI-GA researchers need to be in constant communication with AI safety researchers to help inform them. Ideally, AI-GA researchers should conduct safety research themselves in addition to making AI-GA advances.  Each AI-GA scientist must take precautions to try to ensure that AI-GA research is safe and, should it succeed, that it produces AIs whose values are aligned with our own.  

It is fair to ask why should I write this paper if I think AI-GA research is more dangerous, as I am attempting to inform people about it potentially being a faster path to general AI and advocating that more people work on this path. One reason is I believe that, on balance, technological advances produce more benefit than harm. That said, this technology is very different and could prove an exception to the rule. A second reason is because I think society is better off knowing about this path and its potential, including its risks and downsides. We might therefore be better prepared to maximize the positive consequences of the technology while working hard to minimize the risks and negative outcomes. Additionally, I find it hard to imagine that, if this is the fastest path to AI, then society will not pursue it. I struggle to think of powerful technologies humanity has not invented soon after it had the capability to do so. Thus, if it is inevitable, then we should be aware of the risks and begin organizing ourselves in a way to minimize those risks. 
Very intelligent people disagree with my conclusion to make knowledge of this technology public. I respect their opinions and have discussed this issue with them at length. It was not an easy decision for me to make. But ultimately I feel that it is a service to society to make these issues public rather than keep them the secret knowledge of a few experts. 

There is another ethical concern, although many will find it incredible and dismiss it as the realm of fantasy or science fiction. We do not know how physical matter such as atoms can produce feelings and sensations like pain, pleasure, or the taste of chocolate, which philosophers call qualia. While some disagree, I think we have no good reason to believe that qualia will not emerge at some point in artificially intelligent agents once they are complex enough. A simple thought experiment makes the point: imagine if the mimic path enabled us to simulate an entire human brain and body, down to each subatomic particle. It seems likely to me that such a simulation would feel the same sensations as its real-world counterpart. 

Recognizing if and when artificial agents are feeling pain, pleasure, and other qualia that are worthy of our ethical considerations is an important subject that we will have to come to terms with in the future. However, that issue is not specific to the method in which AI is produced, and therefore is not unique to the AI-GA path. There is an AI-GA-specific consideration on this front, however. On Earth, there has been untold amounts of suffering produced in animals en route to the production of general AI. Is it ethical to create algorithms in which such suffering occurs if it is essential, or helpful, to produce AI? Should we ban research into algorithms that create such suffering in order to focus energy on creating AI-GAs that do not involve suffering? How do we balance the benefits to humans and the planet of having general AI vs.\ the suffering of virtual agents? These are all questions we will have to deal with as research progresses on AI-GAs. They are related to the general question of ethics for artificial agents, but have unique dimensions worthy of specific consideration. 

Some of these ideas will seem fantastical to many researchers. In fact, it is risky for my career to raise them. However, I feel obligated to let society and our community know that I consider some of these seemingly fantastical outcomes possible enough to merit consideration. For example, even if there is a small chance that we create dangerous AI or untold suffering, the costs are so great that we should discuss that possibility. As an analogy, if there were a 1\% chance that a civilization-ending asteroid could hit Earth in a decade or ten, we would be foolish not to begin discussing how to track it and prevent that catastrophe.  

We should keep in mind the grandeur of the task we are discussing, which is nothing short than the creation of an artificial intelligence smarter than humans. If we succeed, we arguably have also created life itself, by some definitions. We do not know if that intelligence will feel. We do not know what its values might be. We do not know what its intentions towards us may be. We might have an educated guess, but any student of history would recognize that it would be the height of hubris to assume we know with certainty exactly what general AI will be like. Thus, it is important to encourage, instead of silence, a discussion of the risks and ethical implications of creating general artificial intelligence. 

\section{Conclusions}

In this essay I described three paths to producing general artificial intelligence. The `mimic path' builds as many biological details of human brains as possible into computational models, and is pursued largely by neuroscientists, computational neuroscientists, and cognitive scientists. The mimic path is unlikely to be the fastest path to general AI because it attempts to simulate all of the detail of biological brains irrespective of whether they can be ignored or abstracted by different, more efficient, machinery.  

The `manual path' is what most of the machine learning community is currently committed to. It involves two phases. In Phase 1, we identify each of the building blocks necessary to create a complex thinking machine. This is the phase we are currently in: most papers either introduce new candidate building blocks or improvements to previously proposed building blocks. It is unclear how long it might take to identify all of the correct building blocks, including the right variants of each one. The manual path implicitly assumes a Phase 2, where we will undertake the Herculean task of figuring out how to combine all of the correct variants of these building blocks into a complex thinking machine. One of the goals of this essay is simply to make explicit the path most of machine learning is committed to and that it implies a second phase that is rarely discussed. Of course, these phases will in practice overlap. There will be teams that increasingly try to combine existing building blocks while other teams continue to create new building blocks and improve existing ones. I discussed the scientific, engineering, and sociological pros and cons of this manual path to creating general AI. 

I also described an alternative path to AI: creating general AI-generating algorithms, or AI-GAs. This path involves Three Pillars: meta-learning architectures, meta-learning algorithms, and automatically generating effective learning environments. As with the other paths, there are advantages and disadvantages to this approach. A major con is that AI-GAs will require a lot of computation, and therefore may not be practical in time to be the first path to produce general AI. However, AI-GA's ability to benefit more readily from exponential improvements in the availability of compute may mean that it surpasses the manual path before the manual path succeeds. A reason to believe that the AI-GA path may be the fastest to produce general AI is in line with the longstanding trend in machine learning that hand-coded solutions are ultimately surpassed by learning-based solutions as the availability of computation and data increase over time. Additionally, the AI-GA path may win because it does not require the Herculean Phase 2 of the manual path and all of its scientific, engineering, and sociological challenges. Additional benefits of AI-GA research are that fewer people are working on it, making it an exciting, unexplored research frontier. 

All three paths are worthwhile scientific grand challenges. That said, society should increase its investment in the AI-GA path. There are entire fields and billions of dollars devoted to the mimic path. Similarly, most of the machine learning community is pursuing the manual path, including billions of dollars in government and industry funding. Relative to these levels of investment, there is little research and investment in the AI-GA path. While still small relative to the manual path, there has been a recent surge of interest in Pillar 1 (meta-learning architectures) and Pillar 2 (meta-learning algorithms). However, there is little work on Pillar 3, and no work to date on attempting to combine the Three Pillars. Since the AI-GA path might be the fastest path to producing general AI, then society should substantially increase its investment in AI-GA research. Even if one believes the AI-GA path has a 1\%-5\% of being the first to produce general AI, then we should allocate corresponding resources into the field to catalyze its progress. That, of course, assumes we conclude that the benefits of potentially producing general AI faster outweigh the risks of producing it via AI-GAs, which I ultimately do. At a minimum, I hope this paper motivates a discussion on these questions. While there is great uncertainty about which path will ultimately produce general AI first, I think there is little uncertainty that we are underinvesting in a promising area of machine learning research.

Finally, this essay has discussed many of the interesting consequences of building general AI that are unique to producing general AI via AI-GAs. One benefit is being able to produce a large diversity of different types of intelligent beings, and thus accelerating our ability to understand intelligence in general and all its potential manifestations. Doing so may also better help us understand our own single instance of intelligence, much as traveling the world is necessary to truly understand one's hometown. Each different intelligence produced by an AI-GA could also create entire alien histories and cultures from which we can learn from. Downsides unique to AI-GAs were also discussed, including that it might make the sudden, unanticipated production of AI more likely, that it might make producing dangerous forms of AI more likely, and that it may create untold suffering in virtual agents. While I offered my own views on these issues and how I weigh the positives and negatives of this technology for the purpose of deciding whether we should pursue it, a main goal of mine is to motivate others to discuss these important issues. 

My overarching goal in this essay is not to argue that one path to general AI is likely to be better or faster. Instead, it is to highlight that there is an entirely different path to producing general AI that is rarely discussed. Because research in that path is less well known, I briefly summarized some of the research we and others have done to take steps towards creating AI-GAs. I also want to encourage reflection on (1) which path or paths each of us is committed to and why, (2)  the assumptions that underlie each path (3) the reasons why each path might prove faster or slower in the production of general AI, (4) whether society and our community should rebalance our investment in the different paths, and (5) the unique benefits and detriments of each approach, including AI safety and ethics considerations. It is my hope that this essay will improve our collective understanding of the space of possible paths to producing general AI, which is worthwhile for everyone regardless of which path we choose to work on. I also hope this essay highlights that there is a relatively unexplored path that may turn out to be the fastest path in the greatest scientific quest in human history. I find that extremely exciting, and hope to inspire others in the community to join the ranks of those working on it.

\section*{Acknowledgements}
My foremost thanks go to Ken Stanley and Joel Lehman, whose lifetime of excellent, creative work greatly informed, influenced, and inspired my thinking on this subject. For helpful discussions on the ideas in this manuscript and/or comments on the manuscript, I thank both of them and Peter Dayan, Zoubin Ghahramani, Ashley Edwards, Felipe Petroski-Such, Vashisht Madhavan, Joost Huizinga, Adrien Ecoffet, Rui Wang, Fritz Obermeyer, Martin Jankowiak, Miles Brundage, Jack Clark, Tegan Maharaj, David Krueger, and all the members of Uber AI Labs. Jeff Clune was supported by an NSF CAREER award (CAREER: 1453549)

\bibliographystyle{plainnat}
\bibliography{JeffCluneBib}

\end{document}